\documentclass{article}
\pdfoutput=1
\usepackage[utf8]{inputenc}
\usepackage{jheppub} 
\usepackage{xcolor,graphicx,epsfig,psfrag,amsmath,empheq}
\usepackage{bm}
\usepackage{xspace}
\usepackage{comment}
\usepackage{caption}
\usepackage{subcaption}
\usepackage{algorithm}
\usepackage{algpseudocode}

\begin{document}

\title{Robust and Provably Monotonic Networks}
\author{Ouail Kitouni, Niklas Nolte, and Mike Williams}

\affiliation{NSF AI Institute for Artificial Intelligence and Fundamental Interactions \\
Laboratory for Nuclear Science, MIT, Cambridge, MA, USA}

\emailAdd{
kitouni@mit.edu,
nnolte@mit.edu,
mwill@mit.edu}

\abstract{
The Lipschitz constant of the map between the input and output space represented by a neural network is a natural metric for assessing the robustness of the model. 
We present a new method to constrain the Lipschitz constant of dense deep learning models that can also be generalized to other architectures. The method relies on a simple weight normalization scheme during training that ensures the Lipschitz constant of every layer is below an upper limit specified by the analyst.
A simple monotonic residual connection can then be used to make the model monotonic in any subset of its inputs, which is useful in scenarios where domain knowledge dictates such dependence. Examples can be found in algorithmic fairness requirements or, as presented here, in the classification of the decays of subatomic particles produced at the CERN Large Hadron Collider. 
Our normalization is minimally constraining and allows the underlying architecture to maintain higher expressiveness compared to other techniques which aim to either control the Lipschitz constant of the model or ensure its monotonicity.
We show how the algorithm was used to train a powerful, robust, and interpretable discriminator for heavy-flavor-quark decays, which has been adopted for use as the primary data-selection algorithm in the LHCb real-time data-processing system in the current LHC data-taking period known as Run~3.
In addition, our algorithm has also achieved state-of-the-art performance on benchmarks in medicine, finance, and other applications. 
}

\notoc

\maketitle

\section{Introduction}

The sensor arrays of the LHC experiments produce over 100 TB/s of data, more than a zettabyte per year. 
After drastic data-reduction performed by custom-built read-out electronics, the annual data volumes are still $O(100)$ exabytes, which cannot be stored indefinitely. 
Therefore, each experiment processes the data in real-time and decides whether each proton-proton collision event should remain persistent or be discarded permanently, referred to as {\em triggering} in particle physics.
Trigger classification algorithms must be designed to minimize the impact of effects like experimental instabilities that occur during data taking---and deficiencies in simulated training samples. (If we knew all of the physics required to produce perfect training samples, there would be no point in performing the experiment.) 
The need for increasingly complex discriminators for the LHCb trigger system~\cite{LHCb-DP-2012-004,LHCb-DP-2019-001} calls for the use of expressive models which are both robust and interpretable. Here we present an architecture based on a novel weight normalization technique that achieves both of these requirements. 

\paragraph{Robustness} A natural way of ensuring the robustness of a model is to constrain the Lipschitz constant of the function it represents, 
defined such that for every pair of points on the graph of the function, the absolute value of the slope of the line connecting them is not greater than the Lipschitz constant.
To this end, we developed a new architecture whose Lipschitz constant is constrained by design using a novel layer-wise normalization which allows the architecture to be more expressive than the current state-of-the-art with more stable and faster training.

\paragraph{Interpretability} An important inductive bias in particle detection at the LHC is the idea that particular collision events are more {\em interesting} if they are outliers, {\em e.g.}, possible evidence of a particle produced with a longer-than-expected (given known physics) lifetime would definitely warrant further detailed study. The problem is that outliers are often caused by experimental artifacts or imperfections, which are included and labeled as background in training; whereas the set of all possible {\em interesting} outliers is not possible to construct {\em a priori}, thus not included in the training process. 
This problem is immediately solved if {\em outliers are better} is implemented directly using an expressive monotonic architecture. Some work was done in this regard \cite{liu2020certified, you2017deepLattice, nips93monotonic} but most implementations are either not expressive enough or provide no guarantees. 
We present Monotonic Lipschitz Networks which overcome both of these problems by building an architecture that is monotonic in any subset of the inputs by design, while keeping the constraints minimal such that it still offers significantly better expressiveness compared to current methods.

\section{Monotonic Lipschitz Networks}
\label{sec:LipNN}
The goal is to develop a neural network architecture representing a scalar-valued function
\begin{align}
f(\bm{x}) : \mathbb{R}^n \rightarrow \mathbb{R}
\end{align}
that is provably monotonic in any subset of inputs and whose gradient (with respect to its inputs) has a constrained magnitude in any particular direction. In an experimental setting, this latter property is a measure of robustness to small changes in experimental conditions or to small deficiencies in the training samples.

Constraints with respect to a particular $L_p$ metric will be denoted as Lip$^p$.
We start with a model $g(\bm{x})$ that is Lip$^1$ with Lipschitz constant $\lambda$ if $\forall \, \bm{x}, \bm{y} \in \mathbb{R}^n$  (we show below how to train such a model)
\begin{align} \label{sigma_lipschitz}
|g(\bm{x}) - g(\bm{y})| &\leq \lambda\|\bm{x} - \bm{y}\|_{1} \,.
\end{align}
The choice of 1-norm is crucial because it allows a well defined maximum directional derivative for each input regardless of the gradient direction. This has the convenient side effect that we can tune the robustness requirement for each input individually.
Note that rescaling the inputs $x_i$ allows for $\lambda$ directional dependence. 

\subsection{Enforcing Monotonicity} 
Assuming we have trained a model that satisfies \eqref{sigma_lipschitz}, 
we can make
an architecture with built-in monotonicity by adding a term that is linear (or has
gradient $\lambda$) in each direction in which we want to be monotonic:
\begin{equation} \label{eqn:monotonic_function}
f(\bm{x}) = g(\bm{x}) + \lambda\sum_{i \in I} x_i,
\end{equation} 
where $I$ denotes the set of indices of the input features for which we would like to be monotonic. This residual connection enforces monotonicity:
\begin{align}
\frac{\partial f}{\partial x_i} = \frac{\partial g}{\partial x_i} + \lambda \geq 0 \quad \forall \, i \in I \,.
\end{align}
Note that the construction presented here only works with Lip$^1$ constraints
as Lip$^{p\neq1}$ functions introduce dependencies between the partial derivatives.
In addition, we stress that  monotonicity is defined via partial derivatives. The value of $f$ is guaranteed to increase when $x_i$ is increased while keeping all $x_{\neq i}$ constant. It is therefore advisable to look out for ill defined edge cases. For instance, let $x_2 \equiv -x_1$ in the training data and define $I = \{1,2\}$. This is incompatible with the architecture and produces unwanted results unless $\lambda=0$ for both $x_1$ and $x_2$ (otherwise the problem is ill posed).

To the best of our knowledge, the only use of residual connections in the literature when trying
to learn monotonic functions is in the context of invertible ResNets \cite{behrmann2019invertible}.
Instead, the state-of-the-art approach for learning monotonic functions involves penalizing negative
gradients in the loss, then certifying the final model is monotonic,
rather than enforcing it in the architecture ({\em e.g.}\ in \cite{liu2020certified}).

\subsection{Enforcing Lipschitz Constraints} %

Ideally, the construction $g(\bm{x})$ should be a universal approximator of Lip$^1$ functions.
Here, we discuss possible architectures for this task.

\paragraph{\texorpdfstring{Lip$^1$}{Lip-1} constrained models}
Fully connected networks can be Lipschitz bounded by constraining the matrix
norm of all weights \cite{p_norm_paper, spectral2018}.
Given the fully connected network with activation $\sigma$
\begin{align}
\label{eqn:gx}
g(\bm{x}) = W^m \sigma ( W^{m-1} \sigma ( ... \sigma(W^1 \bm{x} + b^1) ... )
             + b^{m-1} ) + b^m,
\end{align}
where $W^m$ is the weight matrix of layer $m$, $g(\bm{x})$ satisfies Eq.~\eqref{sigma_lipschitz} if
\begin{align} \label{eqn:matrix_norm_sigma}
\prod_{i=0}^m \|W^i\|_1 \leq \lambda
\end{align}
and $\sigma$ has a Lipschitz constant less than or equal to 1.
There are multiple ways to enforce Eq.~\eqref{eqn:matrix_norm_sigma}.
Two possibilities that involve scaling by the operator norm of the weight matrix \cite{p_norm_paper} are
\begin{align}
\label{eqn:norm}
W^i \rightarrow W'^i = \lambda^{1/m} \frac{W^i}{\text{max}(1, \| W^i \|_1)}
\qquad {\rm or} \qquad  W^i \rightarrow W'^i = \frac{W^i}{\text{max}(1, \lambda^{-1/m} \cdot \| W^i \|_1)}\,.
\end{align}
In our studies thus far, the latter variant seems to train slightly better.
However, in some cases it might be useful to use the former to avoid the scale imbalance between the neural network's output and the residual connection used to induce monotonicity.

In order to satisfy Eq.~\eqref{eqn:matrix_norm_sigma}, it is not necessary to divide
the entire matrix by its 1-norm. It is sufficient to ensure that
the absolute sum over each column is constrained:
\begin{align} \label{eq:column_wise_norm}
W^i \rightarrow W'^i =W^i\mathrm{diag} \left(\frac{1}{\text{max}\left(1, \lambda^{-1/m} \sum_j |W^i_{jk}|\right)}\right)\,.
\end{align}
This novel normalization scheme tends to give even better training results in practice. 
While Eq.~\eqref{eq:column_wise_norm} is not suitable as a general-purpose scheme, {\em e.g.}\ it would not work in convolutional networks, its performance in training in our analysis motivates further study of this approach in future work. 

The constraints in Eqs.~\eqref{eqn:norm} and \eqref{eq:column_wise_norm} can be applied in different ways. For example, one could normalize the weights directly before each call such that the induced gradients are propagated through the network like in \cite{spectral2018}. While one could come up with toy examples for which propagating the gradients in this way hurts training, it appears that this approach is what usually is implemented for spectral norm \cite{spectral2018} in PyTorch and TensorFlow. Alternatively, the constraint could be applied by projecting any infeasible parameter values back into the set of feasible matrices after each gradient update as in Algorithm 2 of \cite{p_norm_paper}. 
Algorithm~\ref{alg:lipnn} summarizes our approach.

\begin{algorithm}
\caption{Training with enforced Lipschitz constraint using weight-norming}\label{alg:lipnn}
\begin{algorithmic}
\Require $\{\mathbb{D}_i\}_{i=1}^n$, a collection of $n$ training batches.
\Require $w$, the non-normalized weight parameter at some layer.
\Comment{These are the optimized leaf parameters}
\Require \textbf{{Norm}}, the function used to normalize the weights, {\em e.g.}\ as given by Eq.~\eqref{eq:column_wise_norm}.
\Require \textbf{{Cost}}, the loss computed using a neural network with weight parameters $\hat w$ on a given batch.\\
\State $\hat{w} \gets \mathrm{\textbf{Norm}}(w) $\Comment{This is the weight used in the neural network matrix multiplication}
\While{not converged}
\For{$i$ from 1 to $n$}
\State $L \gets \mathrm{\textbf{Cost}}(\mathbb{D}_i, \hat{w})$
\State $w \gets w - \nabla_w \hat{w} \cdot \nabla_{\hat{w}} L$
\State $\hat{w} \gets  \mathrm{\textbf{Norm}}(w)$
\EndFor
\EndWhile
\Comment{ At inference time, only $\hat{w}$ is used.}
\end{algorithmic}
\end{algorithm}

\paragraph{Preserving expressive power}
Some Lipschitz network architectures ({\em e.g.}\ \cite{spectral2018}) tend to overconstrain the model in the sense that these architectures
cannot fit all functions $\lambda$-Lip$^1$ due to {\em gradient attenuation}.
For many problems this is a rather theoretical issue.
However, it becomes a practical problem for the monotonic architecture since
it often works on the edges of its constraints, for instance when
partial derivatives close to zero are required.
The authors of \cite{huster2018} showed that ReLU networks are unable to fit the
function $f(x) = |x|$ if the layers are norm-constrained with $\lambda = 1$.
The reason lies in fact that ReLU, and most other commonly used
activations, do not have unit gradient with respect to the inputs over their entire domain.

While element-wise activations like ReLU cannot have unit gradient over the whole domain without being exactly linear, the authors of \cite{sorting2019} explore
activations that introduce nonlinearities by reordering elements of the input vector. 
They propose the following activation function: 
\begin{align}
\sigma = {\rm \textbf{GroupSort}}, 
\end{align}
which sorts its inputs in chunks (groups) of a fixed size.
This operation has gradient $1$ with respect to every input
and gives architectures constrained
with Eq.~\eqref{eqn:matrix_norm_sigma} increased expressive power. 
In addition, we have found that using this activation function also results in achieving sufficient expressiveness with a small number of weights, making the networks ideal for use in resource-constrained applications.

\section{Example Applications to Simple Models}

Before applying our new architecture to real-time data-processing at the LHC, we first demonstrate that it behaves as expected on some simple toy problems.

\subsection{Robustness to Outliers}

We will demonstrate the robustness that arises from the Lipschitz constraint by making a simple toy regression model to fit to data sampled from a 1-dimensional function with one particularly noisy data point. 
The underlying model that we sample from here has the form 
\begin{align}
\label{eqn:toy2}
y = \sin(x) + \epsilon(x), 
\end{align}
where $\epsilon(x)$ is Gaussian noise with unit variance for one data point and $0.01$ otherwise.
While this toy problem will explicitly show that the Lipschitz network is more robust against outliers than an unconstrained network due to its bounded gradient, it also serves as a proxy for any scenario with deficiencies in the training data. 
{\em N.b.}, due to its bounded gradient a Lipschitz network is also more robust against adversarial attacks and data corruption than an unconstrained model.

\begin{figure}[t!]
    \centering
\includegraphics[width=0.7\linewidth]{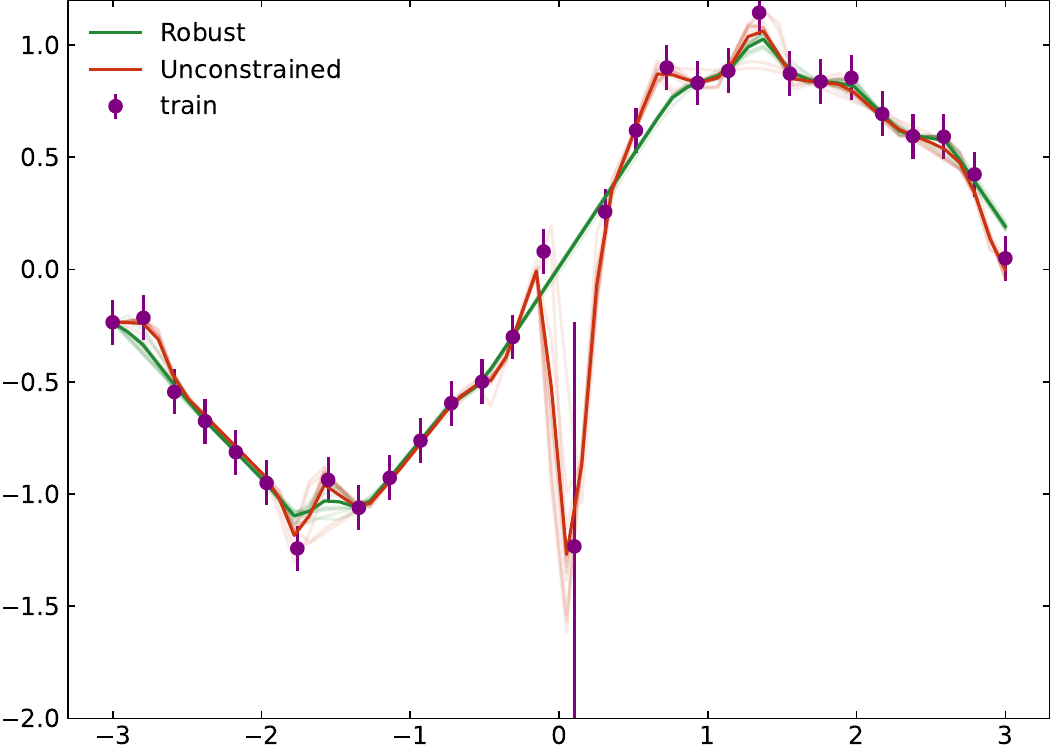}
    \caption{Training robust and unconstrained models using a realization of the toy model in Eq.~\eqref{eqn:toy2}.  
    Each model is trained using 10 random initialization seeds. 
    The dark lines are averages over the seeds, which are each shown as light lines. 
    The unconstrained models exhibit overfitting of the noisy outlier,  whereas the Lipschitz networks are robust. 
    In addition, the Lipschitz constraint produces much smoother models as expected. {\em N.b.}, here we set the Lipschitz constant to be $\lambda = 1$, whereas the slope of the true model is $\cos{x}$. This allows for more variation in the fit model than the true model. In this exercise we assumed that all we know is that the slope is bounded by unity. If we did have more precise {\em a priori} information about the slope, we could easily employ this by rescaling $x$ as discussed in Sec.~\ref{sec:LipNN}. 
    }
    \label{fig:robust}
\end{figure}

Figure~\ref{fig:robust} shows that the unconstrained model  overfits the data as expected, whereas applying our approach from Sec.~\ref{sec:LipNN} does not. 
The Lipschitz model effectively ignores the outlier, since there is no way to accommodate that data point while respecting its built-in gradient bound. 
In addition, we see that the Lipschitz constraint enforces much smoother functions over the full range---the degree of this smoothness determined by us via the chosen Lipschitz constant.  %

\subsection{Monotonic Dependence}

To demonstrate monotonicity, we will make a simple toy regression model to fit to data sampled from the following 1-dimensional function: 
\begin{align}
f(x) = \log(x) + \epsilon(x),
\label{eqn:toy1}
\end{align}
where $\epsilon$ is a Gaussian noise term whose variance is linearly increasing in $x$. 
In this toy model, we will assume that our prior knowledge tells us that the function we are trying to fit must be monotonic, despite the non-monotonic behavior observed due to the noise. 
This situation is ubiquitous in real-world applications of AI/ML, but is especially prevalent in the sciences (see, {\em e.g.}, Sec.~\ref{sec:HLT}).  

\begin{figure}[t!]
    \centering
    \begin{subfigure}[]{0.45\textwidth}
    \includegraphics[width=\linewidth]{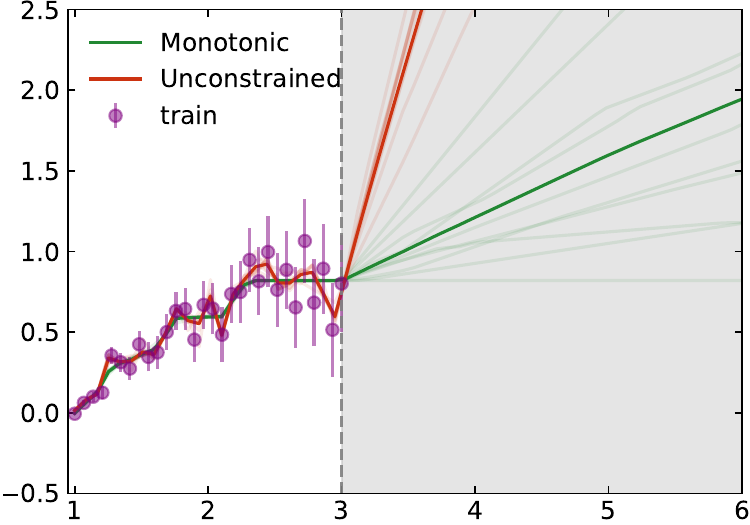}
    \end{subfigure}
    \begin{subfigure}[]{0.45\textwidth}
    \includegraphics[width=\linewidth]{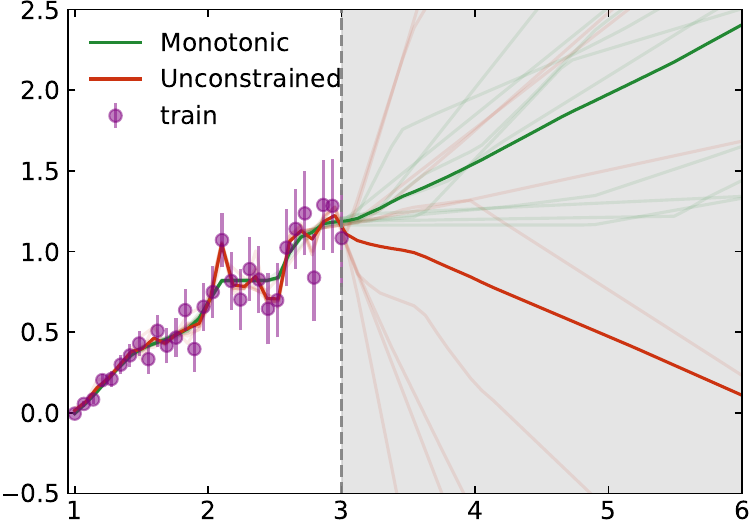}
    \end{subfigure}
    \begin{subfigure}[]{0.45\textwidth}
    \includegraphics[width=\linewidth]{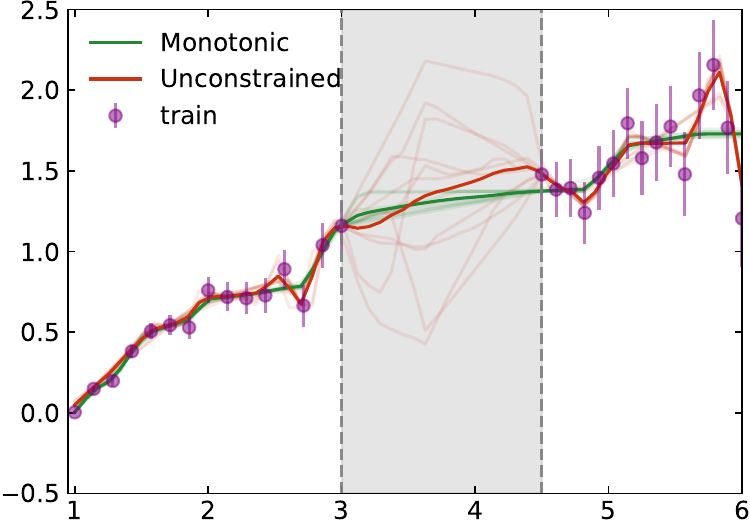}
    \end{subfigure}
    \begin{subfigure}[]{0.45\textwidth}
    \includegraphics[width=\linewidth]{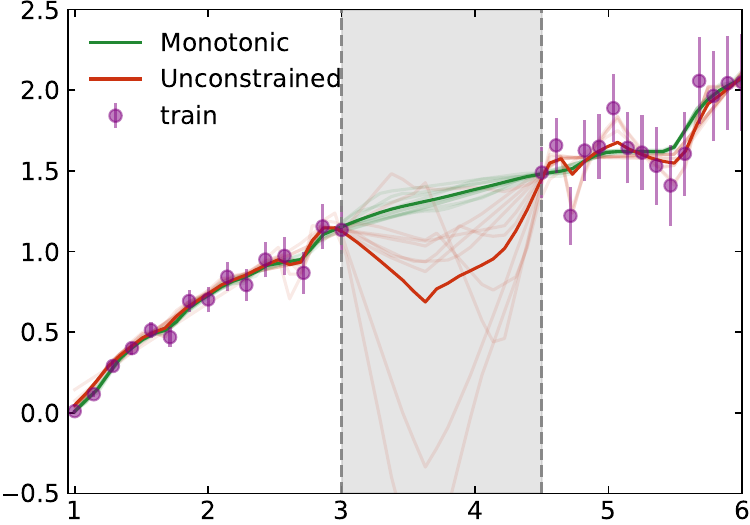}
    \end{subfigure}
    \caption{Training monotonic and unconstrained models using four realizations (purple data points) of the toy model in Eq.~\eqref{eqn:toy1}. 
    The shaded regions represent the (top) extrapolation or (bottom) interpolation regions of interest, where training data are absent. 
    Each panel presents a different realization of the Gaussian noise. 
    Each model is trained using 10 random initialization seeds. %
    The dark lines are averages over the seeds, which are each shown as light lines. 
    The unconstrained models exhibit overfitting of the noise and non-monotonic behavior, and when extrapolating or interpolating into regions where training data were absent, these models exhibit highly undesirable and unpredictable behavior.    
    Conversely, the monotonic Lipschitz models always produce a monotonic function, even in scenarios where the noise is strongly suggestive of non-monotonic behavior. 
    In addition, the Lipschitz constraint produces much smoother models as expected. {\em N.b.}, here we set the Lipschitz constant to be $\lambda = 1$, whereas the slope of the true model is $1/x$. This allows for more variation in the fit model than the true model. In this exercise we assumed that all we know is that the slope is bounded by unity. If we did have more precise {\em a priori} information about the slope, we could easily employ this by rescaling $x$ as discussed in Sec.~\ref{sec:LipNN}. 
        }
    \label{fig:monotonic}
\end{figure}

First, we train standard (unconstrained) neural networks on several different samples drawn from Eq.~\eqref{eqn:toy1}. 
Here, we also consider two generic situations where the training data are missing: one that requires extrapolation beyond the region covered by the training data, and another that requires interpolation between two occupied regions. 
Figure~\ref{fig:monotonic} shows that the unconstrained models overfit the data as expected, resulting in non-monotonic behavior. Furthermore, when extrapolating or interpolating into regions where training data were absent, the unconstrained models exhibit highly undesirable and in some cases unpredictable behavior. 
(This problem is exacerbated in higher dimensions and sparser data.) 
In the case of extrapolation, the behavior of the unconstrained model is largely driven by the noise in the last one or two data points.
The interpolation scenario is less predictable. 

While the overfitting observed here could be reduced by employing some form of strong regularization, such an approach would not (in general) lead to monotonic behavior, nor would it formally bound the gradient.
Applying our approach from Sec.~\ref{sec:LipNN} does both.
Figure~\ref{fig:monotonic} demonstrates that our method always produces a monotonic function, even in the extrapolation scenario where the slope of the noise terms in the last few data points is strongly suggestive of non-monotonic behavior. 
In addition, the Lipschitz constraint produces much smoother models than in the unconstrained case. 
Therefore, we conclude that the monotonicity and Lipschitz constraints do act as strong regularization against fitting random non-monotonic noise as expected.

\subsection{Expressiveness} 

\begin{figure}[h!]
    \centering
    \includegraphics[width=0.8\textwidth]{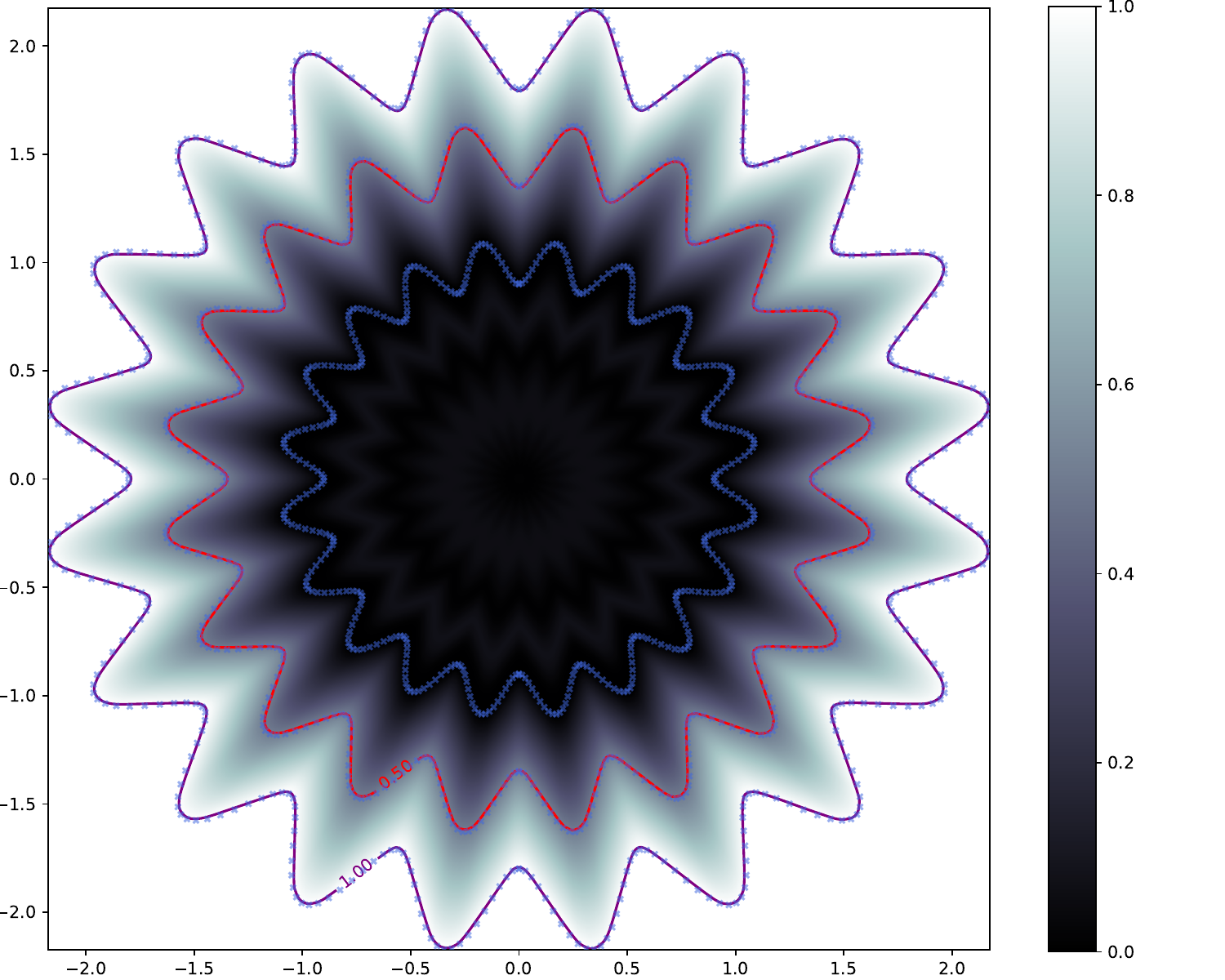}
    \caption{
    Regression example to emulate a complex decision boundary in two dimensions. 
    The training data points are shown in blue (the inner radius is labeled 0, the middle is labeled 0.5, and the outer radius is labeled 1), while the output of the network is shown in color. The contour lines of the network output are shown in purple and red for the values of 1.0 and 0.5, respectively, which properly trace out the curves populated by the outer and middle sets of data points. 
    \label{fig:expressive} 
    }
\end{figure}

GroupSort weight-constrained neural networks can describe arbitrarily complex decision boundaries in classification problems provided the proper objective function is used in training (the usual cross entropy and MSE losses may be sub-optimal for Lipschitz models in some scenarios~\cite{manyfaces}, see Sec.~\ref{sec:improve}). 
Here we will directly regress on a synthetic boundary to emulate a classification problem. The boundary is the perimeter of circle with oscillating radius and is given by
\begin{align}
\partial = \{r + \alpha[\cos\omega\theta,\, \sin\omega\theta]) \,\, |\,\, \theta\in[0, 2\pi]\} \,,
\end{align}
where $r$ and $\alpha$ are chosen to be $1.5$ and $0.18$, respectively.
Figure~\ref{fig:expressive} shows an example where this complicated decision boundary is learned by a Lipschitz network (as defined in Sec.~\ref{sec:LipNN}) trained on the boundary while achieving zero loss, demonstrating the expressiveness that is possible to obtain in these models. 

\section{Example Application: The LHCb inclusive heavy-flavor Run 3 trigger}
\label{sec:HLT}

The architecture presented in Sec.~\ref{sec:LipNN} has been developed with a specific purpose in mind:
The classification of the decays of heavy-flavor particles produced at the Large Hadron Collider, which are bound states that contain a beauty or charm quark that live long enough to travel an observable distance $\mathcal{O}(1\,{\rm cm})$ before decaying.
The dataset used here is built from simulated
proton-proton ($pp$) collisions in the LHCb detector.
Charged particles that survive long enough to traverse the entire detector before decaying are reconstructed and combined pairwise
into decay-vertex (DV) candidates. 

The task concerns discriminating between DV candidates
corresponding to the decays of heavy-flavor particles  versus all other sources of DVs.
The signatures of a heavy-flavor DV are substantial separation from the $pp$ collision point, due to the relatively long heavy-flavor particle lifetimes, and sizable transverse momenta, $p_{\rm T}$, of the component particles, due to the large heavy-flavor particle masses.  
There are three main sources of background DVs.
The first involves DVs formed from particles that originated directly from the $pp$ collision, but where the location of the DV is measured to have non-zero displacement due to resolution effects. 
These DVs will typically have small displacements and small $p_{\rm T}$. 
The second source of background DVs arises due to particles produced in the $pp$ collision interacting with the LHCb detector material, creating new particles at a point in space far from the $pp$ collision point.
Such DVs will have even larger displacement than the signal, but again have smaller $p_{\rm T}$.
The third source involves at least one {\em fake} particle, {\em i.e.}\ a particle inferred from detector information that did not actually exist in the event. 
Since the simplest path through the detector (a straight line) corresponds to the highest possible momentum, DVs involving fake particles can have large $p_{\rm T}$.

In the first decision-making stage of the LHCb trigger, a pre-selection is applied to reject most background DVs, followed by a classifier based on the following four DV features: 
$\sum p_{\rm T}$, the scalar sum of the $p_{\rm T}$ of the two particles that formed the DV;
${\rm min}[\chi^2_{\rm IP}]$, the smaller of the two increases observed when attempting to instead include each component particle into the $pp$-collision vertex fit, which is large when the DV is far from the $pp$ collision point;
the quality of the DV vertex fit;
and the spatial distance between the DV and $pp$-collision locations, relative to their resolutions. 
{\em N.b.}, the threshold required on the classifier response when run in real time during data taking is fixed by the maximum output bandwidth allowed from the first trigger stage.

\begin{figure}
    \centering
     \includegraphics[width=0.99\textwidth]{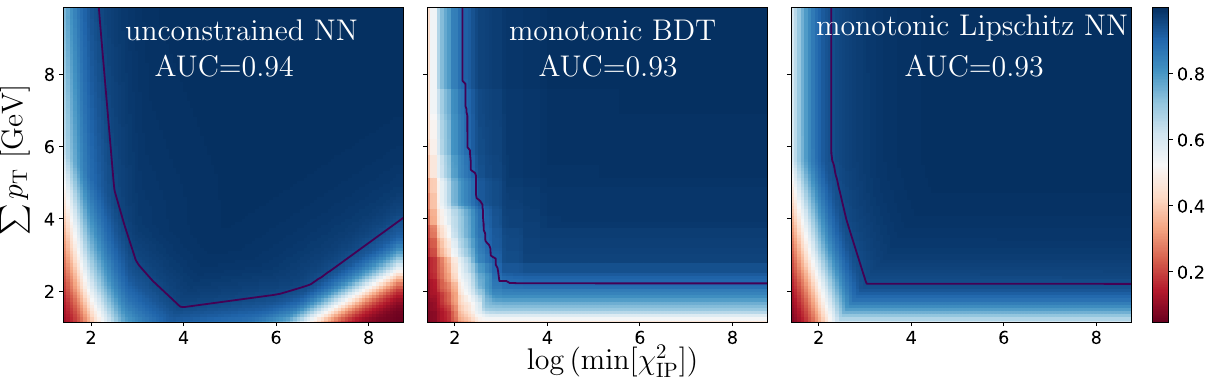} 
    \caption{Simplified version of the LHCb inclusive heavy-flavor trigger problem using only 2 inputs, which permits displaying the response everywhere in the feature space; shown here as a heat map with more signal-like (background-like) regions colored blue (red). The dark solid line shows the decision boundary predicted to give the required output bandwidth in Run~3.}
    \label{fig:heatmaps}
\end{figure}

\begin{figure}
    \centering
     \includegraphics[width=0.99\textwidth]{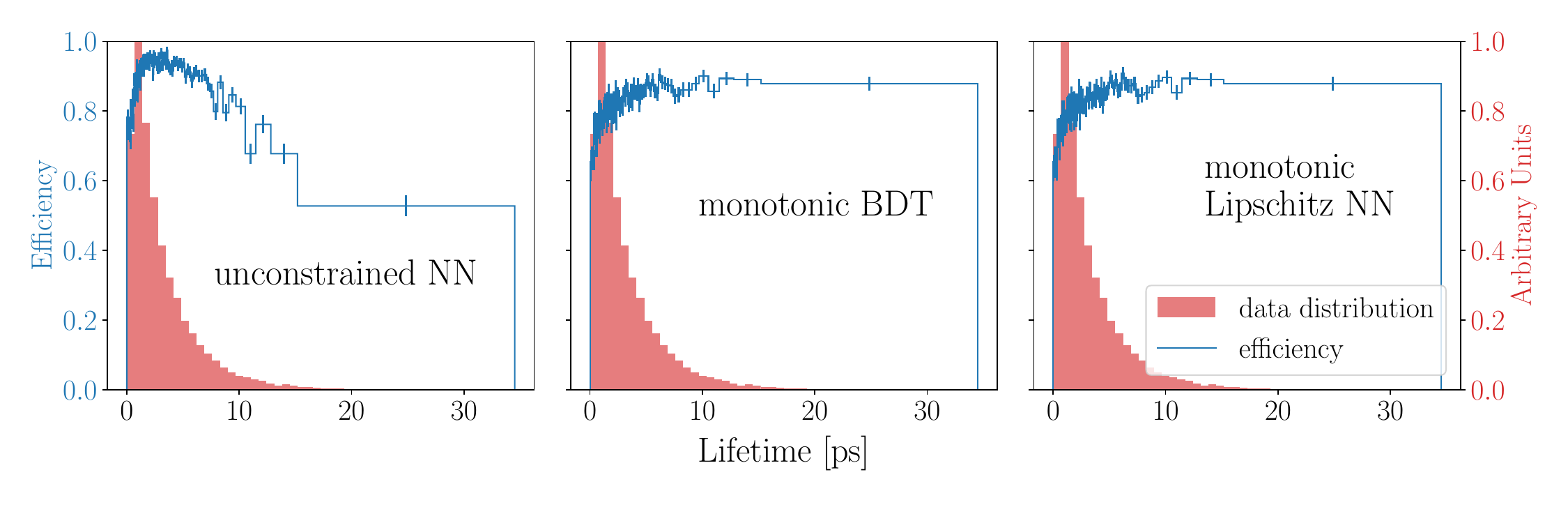} 
    \caption{Efficiency of each model shown in Fig.~\ref{fig:heatmaps} at the expected Run\,3 working point versus the proper lifetime of the decaying heavy-flavor particle selected. The monotonic models produce a nearly uniform efficiency above a few ps at the expense of a few percent lifetime-integrated efficiency. Such a trade off is desirable as explained in the text.}
    \label{fig:efficiency}
\end{figure}

Unfortunately, extremely large values of both displacement and momentum are more common for backgrounds than for heavy-flavor signals. 
For the former, this is easily visualized by considering a simplified  problem using only the two most-powerful inputs, $\sum p_{\rm T}$ and $\chi^2_{\rm IP}$. 
Figure~\ref{fig:heatmaps} (left) shows that an unconstrained neural network learns to reject DVs with increasing larger displacements, corresponding to the lower right corner in the figure. 
Figure~\ref{fig:efficiency} (left) shows that this leads to a dependence of the signal efficiency on the lifetime of the decaying heavy-flavor particle. 
Larger lifetimes are disfavored since few heavy-flavor particles live more than $\mathcal{O}(10\,{\rm ps})$. 
While rejecting DVs with the largest displacements does maximize the integrated signal efficiency in the training sample, this is undesirable because in many cases studying the longest-lived heavy-flavor particles is of more interest than simply collecting the largest decay sample integrated over lifetime (see, {\em e.g.}, \cite{HFLAV:2022pwe}). 
Furthermore, many proposed explanations of dark matter and other types of new physics predict the existence of new particles with similar properties to heavy-flavor particles, but with longer lifetimes~\cite{RF6,Graham:2021ggy}. 
This classifier would reject these particles because it is unaware of our inductive bias that highly displaced DVs are worth selecting in the trigger and studying in more detail later.

Since the LHCb community is generally interested in studying highly displaced DVs for many physics reasons,  
we want to ensure that a larger displacement corresponds to a more signal-like response.
The same goes for DVs with higher $\sum p_{\rm T}$. 
Enforcing a monotonic response in both features is thus a desirable property, especially because it also ensures the desired
behaviour for data points that are outside the boundaries of the training data.
Multiple methods to enforce monotonic behavior in BDTs already exist \cite{auguste2020better}, and Figs.~\ref{fig:heatmaps} (middle) and \ref{fig:efficiency} (middle) show that this approach works here.
However, the jagged decision boundary can cause problems, {\em e.g.}, when measuring the heavy-flavor $p_{\rm T}$ spectrum. 
Specifically, the jagged BDT decision boundary can lead to sharp changes in the selection efficiency. If there is not perfect alignment of where these changes occur with where the interval boundaries of the spectrum are defined, then correcting for the efficiency can be challenging.
Figure~\ref{fig:heatmaps} (right) shows that our novel approach, outlined in Sec.~\ref{sec:LipNN}, successfully produces a smooth and monotonic response, and Fig.~\ref{fig:efficiency} (right) shows that this provides the monotonic lifetime dependence we wanted in the efficiency. 

\begin{figure}[t!]
    \centering
    \includegraphics[width=\textwidth]{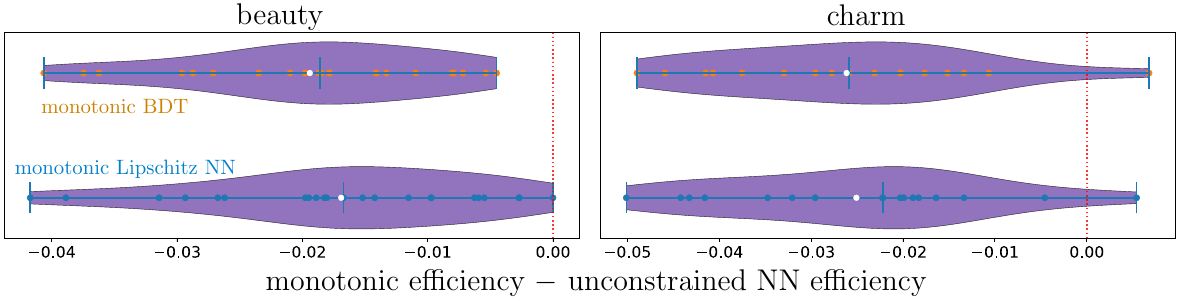}
    \caption{Performance as quantified by the difference in signal efficiency (true positive rate) relative to the unconstrained NN at the expected Run~3 working point for the (left) 24 beauty and (right) 17 charm decays currently being used to benchmark this trigger. Each colored data point shows the change in efficiency for a given decay, while the shaded bands represent the local density of points. The white points show the median values for each set of points.}
    \label{fig:candles}
\end{figure}

Not only does
our architecture guarantee a monotonic response in whatever features the analyst wants, it 
is guaranteed to be robust with respect to
small changes to the inputs as governed by the constrained Lipschitz constant.
Because calibration and resolution effects play a role in obtaining the features during
detector operation, robustness is a necessary requirement for any classification
performed online. 
Downstream analyses of these data depend on their stability.
Figure~\ref{fig:candles} shows that the cost in terms of signal efficiency loss of enforcing monotonicity and robustness is small, even under the unrealistic assumption that the training data were, in fact, perfect. Therefore, the actual cost is likely negligible, while the benefits of the guarantees provided is hard to quantify but immediately obvious to the LHCb collaboration.  
Our algorithm runs in the LHCb trigger software stack and has been chosen to replace Refs.~\cite{BBDT,LHCb-PROC-2015-018} as the primary trigger-selection algorithm used by LHCb in Run~3. 
Due to its guaranteed robustness---and excellent expressiveness even for small networks---this architecture is being explored for other uses within the LHCb trigger system for Run~3, since robustness and monotonicity are ubiquitous inductive biases in experimental particle physics. 

\paragraph{Experiment details} The default LHCb model shown here is a 4-input, 3-layer (width 20) network with GroupSort activation (here, all outputs are sorted), $\lambda \!=\! 2$, constrained using Eq.~\eqref{eq:column_wise_norm}. Inference times in the fully GPU-based LHCb trigger application~\cite{Aaij:2019zbu} are 4 times faster than the Run~3 trigger BDT that was the baseline algorithm before ours was chosen to replace it (the BDT baseline was based on the model used during data taking in Run~2\,\cite{BBDT,LHCb-PROC-2015-018}). We performed $\mathcal{O}(1000)$ runs with different seeds but the differences were negligible, at the level of $\mathcal{O}(0.1\%)$.
For the unconstrained network, we use the same architecture but without the linear term and without the weight constraints during training. The depth and width are the same as used for the monotonic Lipschitz network.
The BDT is a LightGBM \cite{lightgbm} gradient boosted classifier with 1000 base trees and a maximum 25 leaves per tree. Monotonicity is enforced there via the built-in \verb|monotone_constraints| keyword.
Code for the monotonic network implementation of the architecture developed here can be found at \url{https://github.com/niklasnolte/MonotoneNorm}.

\section{Limitations and Potential Improvements} 
\label{sec:improve}

We are working on improving the architecture as follows.
First, common initialization techniques are not optimal for weight-normed networks
\cite{arpit2019initialize}. Simple modifications to weight initialization could likely improve convergence significantly. Second, it appears from empirical investigation that the
networks described in Eq.~\eqref{eqn:matrix_norm_sigma} with GroupSort activation
could be universal approximators, as we have yet to find a function that could not
be approximated well enough with a deep enough network. A proof for universality
is still required and could be developed in the future. Note that universal approximation is indeed proven for a similar architecture that only differs slightly in the normalization scheme, see \cite{sorting2019}.
Neither of these limitations has any practical impact on the example applications discussed in the previous sections. 

In many scenarios, Lipschitz-constrained architectures are considered inferior to unconstrained architectures because of their inability to offer competitive performance on standard benchmarks. This low performance is partly due to the fact that standard losses (such as cross-entropy) are not an adequate proxy of the metric of interest (accuracy) for the Lipschitz-constrained models. At a fundamental level, for any maximally accurate unconstrained classifier $f(x)$ with Lipschitz constant $\lambda$, there exists a Lipschitz 1 classifier that replicates the former's decision boundary, namely, $f(x)/\lambda$. In the following, we will demonstrate a basic toy setting in which a maximally accurate Lipschitz classifier exists but cannot be obtained using standard losses. 

\begin{figure}[h!]
    \centering
\includegraphics[width=\linewidth]{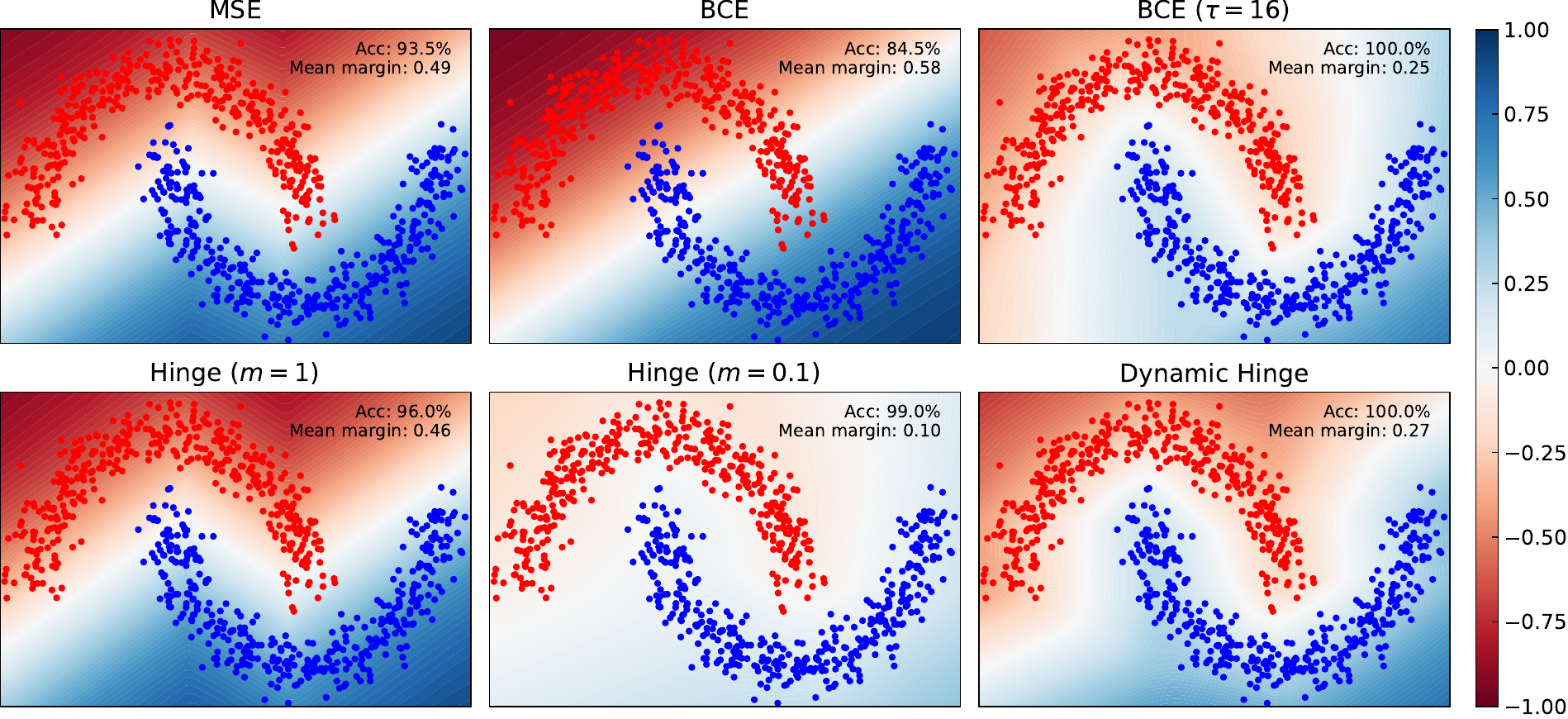}
    \caption{A Lipschitz network trained to classify the Two-Moons dataset using different objects. Ordered from left to right and from top to bottom: Mean Squared Error, Binary Cross Entropy, Binary Cross Entropy with high temperature
    ($\tau=16$), Hinge loss with margin 1,  Hinge loss with margin 0.1, and Hinge with dynamic margin. The network is evaluated on a uniform grid and its output is shown as a heatmap. The average absolute prediction (mean margin) on the validation set is also shown.
    }
    \label{fig:liplosses}
\end{figure}

To understand the effect of the choice of objective function, we train a Lipschitz-constrained model to separate the {\em two-moons} dataset as shown in Fig.~\ref{fig:liplosses}. 
This example is special in that the two samples do not overlap and can be completely separated by a Lipschitz-bounded function; however, that function cannot return the true label values for any data points due to the Lipschitz bound. 
Therefore, a loss function that penalizes any difference of the model output to the true label now faces a misalignment of the optimization target and the actual goal:
While the classification goal is to have high accuracy, {\em i.e.} correct output sign, the optimization target is to minimize deviations of the output from the true label.
This misalignment becomes irrelevant for a function with unbounded Lipschitz constant.
We will show below that for examples such as this there is an important dependence on the objective used and its hyperparameters. 

First, we note that losses with exponential tails (in the sense that they require large weights to reach zero) are in general not suitable for maximizing accuracy. In practice, this can be remedied in cross-entropy by increasing the temperature. Note that cross-entropy with temperature $\tau$ is defined as
\begin{equation}
    \mathcal{L}^{\mathrm{BCE}}_\tau(y, \hat{y}) = \mathcal{L}^{\mathrm{BCE}}(y, \tau \hat{y}),
\end{equation}
where $\mathcal{L}^{\mathrm{BCE}}(y, \hat{y})$ is the usual binary cross-entropy loss on targets $y$ and predictions $\hat{y}$. Following PyTorch conventions, $\hat y$ are logits which will be normalized before computing the negative log-likelihood.

An accurate classification boundary comes at the expense of reduced margins when classes have small separation. A maximally robust accurate classifier will, however, have optimal margins if trained using the appropriate objective. In the case of separable data (\emph{i.e.} when classes have disjoint support), the maximally robust accurate Lipschitz classifier is the signed-distance function (SDF)~\cite{manyfaces} defined, in the binary case as
\begin{equation}
    \mathrm{SDF}(x) = \mathrm{sign}\left [d(C^{+1}, x) - d(C^{-1}, x)\right ] \cdot d(B, x) ,
\end{equation}
where $C^{+1}$ and $C^{-1}$ are the sets of points for which $x$ has label $+1$ and $-1$, respectively, and $B$ is the boundary between classes defined as $B \equiv \left \{x \,\big|\, d(C^{+1}, x) = d(C^{-1}, x)\right \} $. For a closed set $S$, the distance to $x$ is defined as $d(S, x) = \min_{y\in S} d(y, x)$.

A naive objective minimized by the SDF is the hinge loss with margin given by $d(B, x)$. Because we do not have access to the true decision boundary {\em a priori}, as a proxy, we use the following objective: 
\begin{equation}
    \mathcal{L}^{\mathrm{DynamicHinge}}(y, \hat{y}, x) = \mathcal{L}^\mathrm{Hinge}_{\delta(x|y)}(y, \hat{y}),
\end{equation}
where $\delta(x|y) = \frac{d(C^{-y}, x)}{2} $ and $\mathcal{L}^\mathrm{Hinge}$ is defined as 

\begin{equation}
 \mathcal{L}^\mathrm{Hinge}_m(y, \hat{y}) = \max\left(0, m - y\hat{y}\right).
\end{equation}
While this objective produces the highest margins for an accurate classifier, as depicted in Fig.~\ref{fig:liplosses}, it may encounter scalability issues when applied to higher-dimensional problems due to the unavoidable spareness of the training data. There are many possible alternative approaches that could resolve this issue, though this remains an open problem. For lower-dimensional problems with overlapping datasets---as studied in the various examples above and the most common scenario in scientific applications---this non-optimal loss issue does not appear to be relevant.  

Another factor that restricts the perceived expressiveness of Lipschitz architectures is the lack of access to standard techniques that improve convergence in unconstrained networks. For example, batch norm cannot be directly used with Lipschitz architectures. If the variance is too small, it may exceed the Lipschitz bound, and if it is too large, it can reduce the effective Lipschitz constant substantially.

\section{Summary \& Discussion}

The Lipschitz constant of the map between the input and output space represented by a neural network is a natural metric for assessing the robustness of the model. 
We developed a new method to constrain the Lipschitz constant of dense deep learning models that can also be generalized to other architectures. Our method relies on a simple weight normalization scheme during training that ensures the Lipschitz constant of every layer is below an upper limit specified by the analyst.
A simple monotonic residual connection can then be used to make the model monotonic in any subset of its inputs, which is useful in scenarios where domain knowledge dictates such dependence.

Our implementation of Lipschitz constrained networks is minimally constraining compared
to other weight-normed models. 
This allows the underlying architecture to be more
expressive and easier to train while maintaining explicit robustness guarantees.
We showed how the algorithm was used to train a powerful, robust, and interpretable discriminator for heavy-flavor decays in the LHCb trigger system. 
Furthermore, thanks to the expressive capacity of the architecture, we were able to shrink the
number of model parameters to meet the memory and latency requirements of the LHCb trigger, which allows for faster event selection. 
This translates to higher sensitivity
to the elusive physics phenomena we aim to observe. 
Our algorithm has been adopted for use as the primary data-selection algorithm in the LHCb trigger in the current LHC data-taking period known as Run~3.

Given that the desire for robustness and interpretability and benevolent out-of-distribution behavior is ubiquitous when performing experiments, we expect that our architecture could have wide-ranging applications in science. 
In addition, our architecture could also be used
in various applications in which robustness is required such as safety-critical environments
and those which need protection against adversarial attacks. 
Monotonicity is a desirable property in various applications where fairness and safety are a concern. 
There are many scenarios in which models which are not monotonic are unacceptable. For example, 
in Ref.~\cite{ICLR-Mono} we showed that our algorithm achieves state-of-the-art performance on benchmarks in medicine, finance, and other applications with monotonic inductive biases.  
In addition, in Ref.~\cite{NEEMo} we presented a new and interesting direction for the architecture developed here: Estimation of the Wasserstein metric (Earth Mover's Distance) in optimal transport by employing the Kantorovich-Rubinstein duality to enable its use in geometric fitting applications.
Therefore, we expect that our algorithm will have far-reaching impact well beyond  experimental physics.

\section*{Acknowledgement}
The authors would like to thank the LHCb computing and simulation teams for their support and for producing the simulated LHCb samples used to benchmark the performance of RTA software. This work was supported by NSF grants PHY-2019786 (The NSF AI Institute for Artificial Intelligence and Fundamental Interactions, http://iaifi.org/) and OAC-2004645.

\bibliographystyle{JHEP}
\bibliography{bib}

\end{document}